\documentclass[11pt]{article}
\usepackage{coling2016}
\usepackage{times}
\usepackage{latexsym}
\usepackage{amsmath}
\usepackage{graphicx}
\graphicspath{ {images/} }
\usepackage{verbatim}
\usepackage{fancyvrb}
\usepackage{multirow}
\usepackage{multicol}
\usepackage{hhline}
\usepackage{url}
\usepackage{color}
\usepackage[font=footnotesize,labelfont=bf]{caption}
\usepackage{tikz}
\usepackage{standalone}
\usepackage{colortbl}
\usepackage{tabularx}
\usepackage{xcolor}
\usepackage{pgf}
\usepackage{collcell}
\usepackage{framed}

\newcommand\smalltt[1]{\texttt{\small #1}}

\newcommand{\specialcell}[2][c]{%
  \begin{tabular}[#1]{@{}c@{}}#2\end{tabular}}
	
\definecolor{brinkpink}{rgb}{0.98, 0.38, 0.5}

\newlength{\bibitemsep}
\newlength{\bibparskip}
\let\oldthebibliography\thebibliography
\renewcommand\thebibliography[1]{%
  \oldthebibliography{#1}%
  \setlength{\parskip}{\bibitemsep}%
  \setlength{\itemsep}{\bibparskip}%
}

\title{Path-based vs. Distributional Information\\in Recognizing Lexical Semantic Relations}

\author{Vered Shwartz ~~~~~~~~~~~~~~~~~~~~~~~~~~~~~~~~~~~~~~~ Ido Dagan\\
   Computer Science Department, Bar-Ilan University, Ramat-Gan, Israel\\
   {\tt vered1986@gmail.com	~~~~~~	\tt dagan@cs.biu.ac.il }}

\begin{document}

\maketitle

\begin{abstract}
Recognizing various semantic relations between terms is beneficial for many NLP tasks. 
While path-based and distributional information sources are considered complementary for this task, the superior results the latter showed recently suggested that the former's contribution might have become obsolete.
We follow the recent success of an integrated neural method for hypernymy detection \cite{shwartz2016improving} and extend it to recognize multiple relations. The empirical results show that this method is effective in the multiclass setting as well. 
We further show that the path-based information source always contributes to the classification, and analyze the cases in which it mostly complements the distributional information.
\end{abstract}

\section{Introduction} 

\blfootnote{
     
     % final paper: en-us version (to licence, a license)
    
     \hspace{-0.65cm}  % space normally used by the marker
     This work is licenced under a Creative Commons 
     Attribution 4.0 International License.
     License details:\\
     \url{http://creativecommons.org/licenses/by/4.0/}
}

Automated methods to recognize the lexical semantic relation the holds between terms are valuable for NLP applications. Two main information sources are used to recognize such relations: path-based and distributional. Path-based methods consider the \emph{joint} occurrences of the two terms in a given pair in the corpus, where the dependency paths that connect the terms are typically used as features \cite{hearst1992automatic,snow2004learning,nakashole2012patty,riedel2013relation}. Distributional methods are based on the \emph{disjoint} occurrences of each term and have recently become popular using word embeddings \cite{mikolov2013distributed,pennington2014glove}, which provide a distributional representation for each term. These embedding-based methods were reported to perform well on several common datasets \cite{baroni2012entailment,roller2014inclusive}, consistently outperforming other methods \cite{santus2016nine,necsulescu2015reading}.

While these two sources have been considered complementary, recent results suggested that path-based methods have no marginal contribution over the distributional ones. 
Recently, however, \newcite{shwartz2016improving} presented \smalltt{HypeNET}, an integrated path-based and distributional method for hypernymy detection. They showed that a good path representation can provide substantial complementary information to the distributional signal in hypernymy detection, notably improving results on a new dataset.

In this paper we present \smalltt{LexNET}, an extension of \smalltt{HypeNET} that recognizes \emph{multiple} semantic relations. 
We show that this integrated method is indeed effective also in the multiclass setting. In the evaluations reported in this paper, \smalltt{LexNET} performed better than each individual method on several common datasets. Further, it was the best performing system in the semantic relation classification task of the CogALex 2016 shared task \cite{shwartz2016lexnet_shared_task}.

We further assess the contribution of path-based information to semantic relation classification. Even though the distributional source is dominant across most datasets, path-based information always contributed to it. In particular, path-based information seems to better capture the relationship between terms, rather than their individual properties, and can do so even for rare words or senses. Our code and data are available at \url{https://github.com/vered1986/LexNET}.

% Change \texttt style for the network image
\let\oldttdefault=\ttdefault
\renewcommand*\ttdefault{lcmtt}

\begin{figure*}[!h]
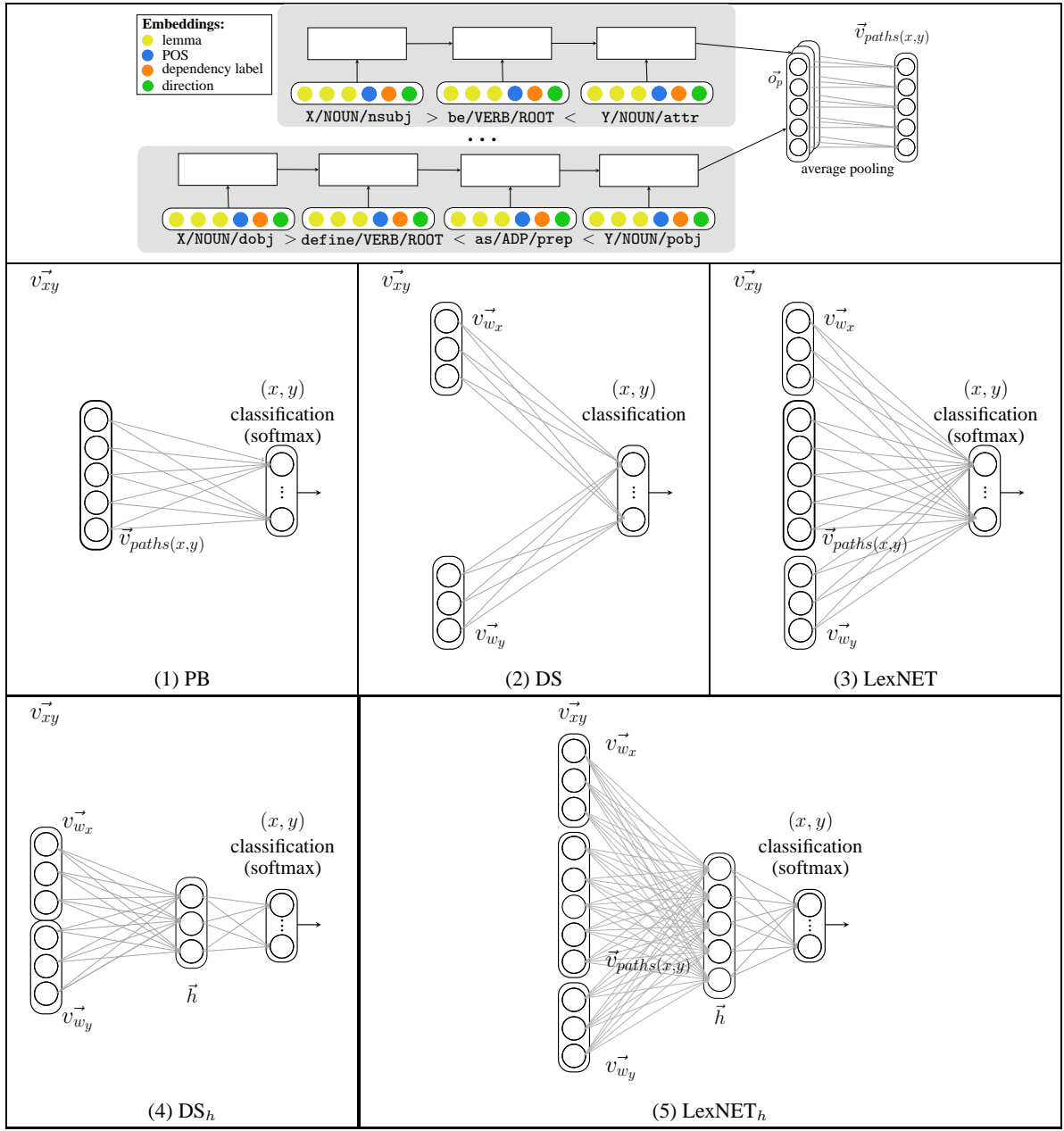
 
\centering
\begin{tabular}{ | c | c | c | }
    \hline
    \multicolumn{3}{|c|}{\resizebox{0.75\linewidth}{!}{\input{images/path_embedding}}} \\ \hline
		\resizebox{0.3\linewidth}{!}{\input{images/pb}} & \resizebox{0.3\linewidth}{!}{\input{images/ds}} & \resizebox{0.3\linewidth}{!}{\input{images/lexnet}} \\ 
		\small(1) PB & \small(2) DS & \small(3) LexNET \\ \hline
		\resizebox{0.3\linewidth}{!}{\input{images/ds_h}} & \multicolumn{2}{|c|}{\resizebox{0.3\linewidth}{!}{\input{images/lexnet_h}}} \\ 
		\small(4) DS$_h$ & \multicolumn{2}{|c|}{\small(5) LexNET$_h$} \\ \hline
\end{tabular}
\vspace*{-5pt}
\caption{Illustrations of classification models. Top row: path-based component. A path is a sequence of edges, and each edge consists of four components: lemma, POS, dependency label and direction. Edge vectors are fed in sequence into the LSTM, resulting in an embedding vector $\vec{o_p}$ for each path. $\vec{v}_{paths(x,y)}$ is the average of $(x, y)$'s path embeddings.}
\label{fig:models}
\vspace*{-10pt}
\end{figure*}

% Revert \texttt style to default
\let\ttdefault=\oldttdefault

\section{Background: HypeNET}
\label{sec:background}

Recently, \newcite{shwartz2016improving} introduced \smalltt{HypeNET}, a hypernymy detection method based on the integration of the best-performing distributional method with a novel neural path representation, improving upon state-of-the-art methods. In \smalltt{HypeNET}, a term-pair $(x, y)$ is represented as a feature vector, consisting of both distributional and path-based features: $\vec{v}_{xy} = [\vec{v}_{w_x}, \vec{v}_{paths(x,y)}, \vec{v}_{w_y}]$, where $\vec{v}_{w_x}$ and $\vec{v}_{w_y}$ are $x$ and $y$'s word embeddings, providing their distributional representation, and $\vec{v}_{paths(x,y)}$ is a vector representing the dependency paths connecting $x$ and $y$ in the corpus. A binary classifier is trained on these vectors, yielding $c = \operatorname{softmax}(W \cdot \vec{v}_{xy})$, predicting hypernymy if $c[1] > 0.5$.

Each dependency path is embedded using an LSTM \cite{hochreiter1997long}, as illustrated in the top row of Figure~\ref{fig:models}. This results in a path vector space in which semantically-similar paths (e.g. \emph{X is defined as Y} and \emph{X is described as Y}) have similar vectors. The vectors of all the paths that connect $x$ and $y$ are averaged to create $\vec{v}_{paths(x,y)}$.

\newcite{shwartz2016improving} showed that this new path representation outperforms prior path-based methods for hypernymy detection, and that the integrated model yields a substantial improvement over each individual model. While HypeNET is designed for detecting hypernymy relations, it seems straightforward to extend it to classify term-pairs simultaneously to multiple semantic relations, as we describe next.

\section{Classification Methods}
\label{sec:methods}

We experiment with several classification models, as illustrated in Figure~\ref{fig:models}:

\paragraph{Path-based} HypeNET's path-based model (\smalltt{PB}) is a binary classifier trained on the path vectors alone: $\vec{v}_{paths(x,y)}$. We adapt the model to classify multiple relations by changing the network softmax output $c$ to a distribution over $k$ target relations, classifying a pair to the highest scoring relation: $r = \operatorname{argmax}_i c[i]$.

\paragraph{Distributional} We train an SVM classifier on the concatenation of $x$ and $y$'s word embeddings $[\vec{v}_{w_x}, \vec{v}_{w_y}]$ \cite{baroni2012entailment} (\smalltt{DS}).\footnote{We experimented also with difference $\vec{v}_{w_x} - \vec{v}_{w_y}$ and other classifiers, but concatenation yielded the best performance.}
\newcite{levy2015supervised} claimed that such a linear classifier is incapable of capturing interactions between $x$ and $y$'s features, and that instead it learns separate properties of $x$ or $y$, e.g. that \emph{y} is a \emph{prototypical hypernym}. To examine the effect of non-linear expressive power on the model, we experiment with a neural network with a single hidden layer trained on $[\vec{v}_{w_x}, \vec{v}_{w_y}]$ (\smalltt{DS$_h$}).\footnote{This was previously done by \newcite{bowman2014learning}, with promising results, but on a small artificial vocabulary.}

\paragraph{Integrated} We similarly adapt the HypeNET integrated model to classify multiple semantic relations (\smalltt{LexNET}). Based on the same motivation of \smalltt{DS$_h$}, we also experiment with a version of the network with a hidden layer (\smalltt{LexNET$_h$}), re-defining $c = \operatorname{softmax}(W_2 \cdot \vec{h} + b_2)$, where $\vec{h} = \operatorname{tanh}(W_1 \cdot \vec{v}_{xy} + b_1)$ is the hidden layer. The technical details of our network are identical to \newcite{shwartz2016improving}.

\section{Datasets}
\label{sec:datasets}

\begin{table}[!t]
\center
\small
\begin{tabular}{ | c | c | c | }
    \hline
    \textbf{dataset} & \specialcell{\textbf{dataset relations}} & \textbf{\#instances} \\ \hline
		\multirow{1}{*}{\textbf{K\&H+N}} & hypernym, meroynym, co-hyponym, random & \multirow{1}{*}{57,509} \\ \hline
		\multirow{1}{*}{\textbf{BLESS}} & hypernym, meroynym, co-hyponym, , event, attribute, random & \multirow{1}{*}{26,546} \\ \hline
		\multirow{1}{*}{\textbf{ROOT09}} & hypernym, co-hyponym, random & \multirow{1}{*}{12,762} \\ \hline
		\multirow{1}{*}{\textbf{EVALution}} & hypernym, meronym, attribute, synonym, antonym, holonym, substance meronym & \multirow{1}{*}{7,378} \\ \hline
\end{tabular}
	\vspace{-5pt}
	\caption{The relation types and number of instances in each dataset, named by their WordNet equivalent where relevant.}
	\label{tab:datasets}
	\vspace{-10pt}
\end{table}

We use four common semantic relation datasets that were created using semantic resources: \smalltt{K\&H+N} \cite{necsulescu2015reading} (an extension to \newcite{kozareva2010semi}), \smalltt{BLESS} \cite{baroni2011we}, \smalltt{EVALution} \cite{santus2015evalution}, and \smalltt{ROOT09} \cite{santus2016nine}. 

Table~\ref{tab:datasets} displays the relation types and number of instances in each dataset. Most dataset relations are parallel to WordNet relations, such as hypernymy (\emph{cat, animal}) and meronymy (\emph{hand, body}), with an additional \emph{random} relation for negative instances. \smalltt{BLESS} contains the \emph{event} and \emph{attribute} relations, connecting a concept with a typical activity/property (e.g. \emph{(alligator, swim)} and \emph{(alligator, aquatic)}). \smalltt{EVALution} contains a richer schema of semantic relations, with some redundancy: it contains both meronymy and holonymy (e.g. for \emph{bicycle} and \emph{wheel}), and the fine-grained substance-holonymy relation. We removed two relations with too few instances: \emph{Entails} and \emph{MemberOf}.

To prevent the lexical memorization effect \cite{levy2015supervised}, \newcite{santus2016nine} added negative switched hyponym-hypernym pairs (e.g. \emph{(apple, animal)}, \emph{(cat, fruit)}) to \smalltt{ROOT09}, which were reported to reduce this effect.

\section{Results}
\label{sec:evaluation}

Like \newcite{shwartz2016improving}, we tuned the methods' hyper-parameters on the validation set of each dataset, and used Wikipedia as the corpus. Table~\ref{tab:results} displays the performance of the different methods on all datasets, in terms of recall, precision and $F_1$.\footnote{Additional evaluation of the method is available in our CogALex 2016 shared task submission \cite{shwartz2016lexnet_shared_task}.}

Our first empirical finding is that Shwartz et al.'s \shortcite{shwartz2016improving} algorithm is effective in the multiclass setting as well (\smalltt{LexNET}).
The only dataset on which performance is mediocre is \smalltt{EVALution}, which seems to be inherently harder for all methods, due to its large number of relations and small size.
The performance differences between \smalltt{LexNET} and \smalltt{DS} are statistically significant on all datasets with p-value of 0.01 (paired t-test). The performance differences between \smalltt{LexNET} and \smalltt{DS$_h$} are statistically significant on \smalltt{BLESS} and \smalltt{ROOT09} with p-value of 0.01, and on \smalltt{EVALution} with p-value of 0.05.

\smalltt{DS$_h$} consistently improves upon \smalltt{DS}. The hidden layer seems to enable interactions between $x$ and $y$'s features, which is especially noticed in \smalltt{ROOT09}, where the hypernymy $F_1$ score in particular rose from 0.25 to 0.45. Nevertheless, we did not observe a similar behavior in \smalltt{LexNET$_h$}, which worked similarly or slightly worse than \smalltt{LexNET}. It is possible that the contributions of the hidden layer and the path-based source over the distributional signal are redundant.\footnote{We also tried adding a hidden layer only over the distributional features of \smalltt{LexNET}, but it did not improve performance.}
It may also be that the larger number of parameters in \smalltt{LexNET$_h$} prevents convergence to the optimal values given the modest amount of training data, stressing the need for large-scale datasets that will benefit training neural methods.

\begin{table*}[!t]
\center
\small
\hspace*{-10pt}
\begin{tabular}{ c | c | c | c || c | c | c || c | c | c || c | c | c | }
    \hhline{~------------}
		& \multicolumn{3}{c||}{\textbf{K\&H+N}} & \multicolumn{3}{c||}{\textbf{BLESS}} & \multicolumn{3}{c||}{\textbf{ROOT09}} & \multicolumn{3}{c|}{\textbf{EVALution}} \\ \hline
    \multicolumn{1}{|c|}{\textbf{method}} & \textbf{P} & \textbf{R} & \boldmath $F_1$ & \textbf{P} & \textbf{R} & \boldmath $F_1$ 
											& \textbf{P} & \textbf{R} & \boldmath $F_1$ & \textbf{P} & \textbf{R} & \boldmath $F_1$\\ \hline
		\multicolumn{1}{|l|}{PB} & 0.713 & 0.604 & 0.55 & 0.759 & 0.756 & 0.755 & 0.788 & 0.789 & 0.788 & 0.53 & 0.537 & \multicolumn{1}{c|}{0.503} \\ \hline
		\multicolumn{1}{|l|}{DS} & 0.909 & 0.906 & 0.904 & 0.811 & 0.812 & 0.811 & 0.636 & 0.675 & 0.646 & 0.531 & 0.544 & \multicolumn{1}{c|}{0.525} \\ \hline
		\multicolumn{1}{|l|}{DS$_h$} & 0.983 & 0.984 & 0.983 & 0.891 & 0.889 & 0.889 & 0.712 & 0.721 & 0.716 & 0.57 & 0.573 & \multicolumn{1}{c|}{0.571} \\ \hline
		\multicolumn{1}{|l|}{LexNET} & 0.985 & 0.986 & 0.985 & 0.894 & \textbf{0.893} & \textbf{0.893} & 
		\textbf{0.813} & 0.814 & 0.813 & \textbf{0.601} & \textbf{0.607} & \multicolumn{1}{c|}{\textbf{0.6}} \\ \hline
		\multicolumn{1}{|l|}{LexNET$_h$} & 0.984 & 0.985 & 0.984 & \textbf{0.895} & 0.892 & \textbf{0.893} & 
		0.812 & \textbf{0.816} & \textbf{0.814} & 0.589 & 0.587 & \multicolumn{1}{c|}{0.583} \\ \hline
	\end{tabular}
	\vspace{-5pt}
	\caption{Performance scores (precision, recall and $F_1$) of each individual approach and the integrated models. 
	To compute the metrics we used scikit-learn \cite{scikit-learn} with the ``averaged'' setup, which computes the metrics for each relation, and reports their average, weighted by support (the number of true instances for each relation). Note that it can result in an $F_1$ score that is not the harmonic mean of precision and recall.}
	\label{tab:results}
	\vspace{-8pt}
\end{table*}

\section{Analysis}
\label{sec:analysis}

Table~\ref{tab:results} demonstrates that the distributional source is dominant across most datasets, with \smalltt{DS} performing better than \smalltt{PB}. Although by design \smalltt{DS} does not consider the relation between $x$ and $y$, but rather learns properties of $x$ or $y$, it performs well on \smalltt{BLESS} and \smalltt{K\&H+N}. \smalltt{DS$_h$} further manages to capture relations at the distributional level, leaving the path-based source little room for improvement on these two datasets.

On \smalltt{ROOT09}, on the other hand, \smalltt{DS} achieved the lowest performance. Our analysis reveals that this is due to the switched hypernym pairs, which drastically hurt the ability to memorize individual single words, hence reducing performance. The $F_1$ scores of \smalltt{DS} on this dataset were 0.91 for co-hyponyms but only 0.25 for hypernyms, while \smalltt{PB} scored 0.87 and 0.66 respectively. Moreover, \smalltt{LexNET} gains 10 points over \smalltt{DS$_h$}, suggesting the better capacity of path-based methods to capture relations between terms. 

\subsection{Analysis of Information Sources}
\label{sec:compare_by_approach}

To analyze the contribution of the path-based information source, we examined the term-pairs that were correctly classified by the best performing integrated model (\smalltt{LexNET}/\smalltt{LexNET$_h$}) while being incorrectly classified by DS$_h$. Table~\ref{tab:qualitative} displays the number of such pairs in each dataset, with corresponding term-pair examples. The common errors are detailed below:

\begin{table}[!t]
\center
\small
\begin{tabular}{ | c | c | c | c | c | c | c | }
    \hline
    \textbf{dataset} & \textbf{\#pairs} & \textbf{x} & \textbf{y} & \textbf{gold label} & \textbf{DS$_h$ prediction} & \textbf{possible explanation} \\ \hline
		\multirow{3}{*}{\textbf{K\&H+N}} & \multirow{3}{*}{\textbf{102}} & firefly & car & false & hypo & \emph{(x, car)} frequent label is hypo \\
																					& & racehorse & horse & hypo & false & out of the embeddings vocabulary \\ 
																					& & larvacean & salp & sibl & false & rare terms \emph{larvacean} and \emph{salp} \\ \hline 	                    
		\multirow{3}{*}{\textbf{BLESS}} & \multirow{3}{*}{\textbf{275}} & tanker & ship & hyper & event & \emph{(x, ship)} frequent label is event \\
																				& & squirrel & lie & random & event & \emph{(x, lie)} frequent label is event \\ 
																				& & herring & salt & event & random & non-prototypical relation \\	\hline
		\multirow{3}{*}{\textbf{ROOT09}} & \multirow{3}{*}{\textbf{562}} & toaster & vehicle & RANDOM & HYPER & \emph{(x, vehicle)} frequent label is HYPER \\
																				& & rice & grain & HYPER & RANDOM & \emph{(x, grain)} frequent label is RANDOM \\ 				
																				& & lung & organ & HYPER & COORD & polysemous term \emph{organ} \\ \hline
		\multirow{3}{*}{\textbf{EVALution}} & \multirow{3}{*}{\textbf{235}} & pick & metal & MadeOf & IsA & rare sense of \emph{pick} \\
																				& & abstract & concrete & Antonym & MadeOf & polysemous term \emph{concrete} \\	
																				& & line & thread & Synonym & MadeOf & \emph{(x, thread)}  frequent label is MadeOf \\ \hline 
\end{tabular}
	\vspace{-5pt}
	\caption{The number of term-pairs that were correctly classified by the integrated model while being incorrectly classified by DS$_h$, in each test set, with corresponding examples of such term-pairs.}
	\label{tab:qualitative}
	\vspace{-10pt}
\end{table}

\paragraph{Lexical Memorization} \smalltt{DS$_h$} often classifies $(x,y)$ term-pairs according to the most frequent relation of one of the terms (usually $y$) in the train set. The error is mostly prominent in \smalltt{ROOT09} (405/562 pairs, 72\%), as a result of the switched hypernym pairs. However, it is not infrequent in other datasets (47\% in \smalltt{BLESS}, 43\% in \smalltt{EVALution} and 34\% in \smalltt{K\&H+N}). As opposed to distributional information, path-based information pertains to both terms in the pair. With such information available, the integrated model succeeds to overcome the most frequent label bias for single words, classifying these pairs correctly.

\paragraph{Non-prototypical Relations} \smalltt{DS$_h$} might fail to recognize non-prototypical relations between terms, i.e. when $y$ is a less-prototypical relatum of $x$, as in \emph{mero:(villa, guest)}, \emph{event:(cherry, pick)}, and \emph{attri:(piano, electric)}. While being overlooked by the distributional methods, these relations are often expressed in joint occurrences in the corpus, allowing the path-based component to capture them.

\paragraph{Rare Terms} The integrated method often managed to classify term-pairs in which at least one of the terms is rare (e.g. \emph{hyper:(mastodon, proboscidean)}), where the distributional method failed. It is a well known shortcoming of path-based methods that they require informative co-occurrences of $x$ and $y$, which are not always available. With that said, thanks to the averaged path representation, \smalltt{PB} can capture the relation between terms even if they only co-occur once within an informative path, while the distributional representation of rare terms is of lower quality. We note that the path-based information of $(x,y)$ is encoded in the vector $\vec{v}_{paths(x,y)}$, which is the averaged vector representation of all paths that connected $x$ and $y$ in the corpus. Unlike other path-based methods in the literature, this representation is indifferent to the total number of paths, and as a result, even a single informative path can lead to successful classification.

\paragraph{Rare Senses} Similarly, the path-based component succeeded to capture relations for rare senses of words where \smalltt{DS$_h$} failed, e.g. \emph{mero:(piano, key), event:(table, draw)}. Distributional representations suffer from insufficient representation of rare senses, while \smalltt{PB} may capture the relation with a single meaningful occurrence of the rare sense with its related term. At the same time, it is less likely for a polysemous term to co-occur, in its non-related senses, with the candidate relatum. For instance, paths connecting \emph{piano} to \emph{key} are likely to correspond to the keyboard sense of \emph{key}, indicating the relation that does hold for this pair with respect to this rare sense. 

Finally, we note that \smalltt{LexNET}, as well as the individual methods, perform poorly on synonyms and antonyms.
The synonymy $F_1$ score in \smalltt{EVALution} was 0.35 in \smalltt{LexNET} and in \smalltt{DS$_h$} and only 0.09 in \smalltt{PB}, reassessing prior findings \cite{mirkin2006integrating} that the path-based approach is weak in recognizing synonyms, which do not tend to co-occur. \smalltt{DS$_h$} performed poorly also on antonyms ($F_1 = 0.54$), which were often mistaken for synonyms, since both tend to occur in the same contexts. It seems worthwhile to try improving the model using insights from prior work on these specific relations \cite{santus2014unsupervised,mohammad2013computing} or additional information sources, like multilingual data \cite{PavlickEtAl-2015:ACL:Semantics}.

\section{Conclusion}

We presented an adaptation to HypeNET \cite{shwartz2016improving} that classifies term-pairs to one of multiple semantic relations. Evaluation on common datasets shows that HypeNET is extensible to the multiclass setting and performs better than each individual method. 

Although the distributional information source is dominant across most datasets, it consistently benefits from path-based information, particularly when finer modeling of inter-term relationship is needed.

Finally, we note that all common datasets were created synthetically using semantic resources, leading to inconsistent behavior of the different methods, depending on the particular distribution of examples in each dataset. This stresses the need to develop ``naturally'' distributed datasets that would be drawn from corpora, while reflecting realistic distributions encountered by semantic applications.

\section*{Acknowledgments}
This work was partially supported by an Intel ICRI-CI grant, the Israel Science Foundation grant 880/12, and the German Research Foundation through the German-Israeli Project Cooperation (DIP, grant DA 1600/1-1).

\bibliography{multiclass}
\bibliographystyle{acl}

\end{document}